\documentclass[a4paper]{svproc}
\usepackage{graphicx}
\usepackage{xcolor}
\usepackage{bm}
\usepackage{amsmath}
\usepackage{amssymb}
\usepackage{mathrsfs}
\usepackage{cite}
\usepackage{subfigure}
\usepackage{tabularx}
\usepackage[normalem]{ulem}
\usepackage[numbers, sectionbib]{natbib}
\usepackage[bookmarks=true]{hyperref}
\usepackage{xurl}

\DeclareMathOperator*{\argmax}{arg\,max}

\begin{document}
\title{
Decision-Making Among Bounded Rational Agents
}
\titlerunning{Decision-Making Among Bounded Rational Agents}
%
\author{
Junhong Xu$^1$ 
\and
Durgakant Pushp \and
Kai Yin \and
Lantao Liu
}
\institute{}
\author{Junhong Xu\inst{1} \and
Durgakant Pushp\inst{1} \and
Kai Yin\inst{2} \and
Lantao Liu\inst{1}}

\authorrunning{J. Xu, D. Pushp, et al.}
%
\institute{Indiana University, Bloomington\inst{1}, IN 47408, USA \\
            Expedia Group\inst{2}\\
\email{\{xu14, dpushp, lantao\}@iu.edu, yinkai1000@gmail.com}\\
}

\maketitle      
\begin{abstract}
When robots share the same workspace with other intelligent agents (e.g., other robots or humans),
they must be able to reason about the behaviors of their neighboring agents while accomplishing the designated tasks. 
In practice, frequently, agents do not exhibit absolutely rational behavior due to their limited computational resources. 
Thus, predicting the optimal agent behaviors is undesirable (because it demands prohibitive computational resources) and undesirable (because the prediction may be wrong).
Motivated by this observation, we remove the assumption of perfectly rational agents and propose incorporating the concept of bounded rationality from an information-theoretic view into the game-theoretic framework.
This allows the robots to reason other agents' sub-optimal behaviors and act accordingly under their computational constraints. 
Specifically, bounded rationality directly models the agent's information processing ability, which is represented as the KL-divergence between nominal and optimized stochastic policies, and the solution to the bounded-optimal policy can be obtained by an efficient importance sampling approach. 
Using both simulated and real-world experiments in multi-robot navigation tasks, we demonstrate that the resulting framework allows the robots to reason about different levels of rational behaviors of other agents and compute a reasonable strategy under its computational constraint. 
\footnote{\small A preliminary version of this work appeared as a poster in {\em 2021 NeurIPS Workshop on Learning and Decision-Making with Strategic Feedback}.\\
The video of the real-world experiments can be found at \url{https://youtu.be/hzCitSSuWiI}. \\
We gratefully acknowledge the support of NSF with grant No. 2006886 and 2047169.}

\keywords{Bounded Rationality, Game Theory, Multi-Robot System}
\end{abstract}
\section{Introduction}
\label{sec:intro}
We consider the problem of generating reasonable decisions for robots in multi-agent environments.
This decision-making problem is complex because each robot's motion trajectory depends on and affects the trajectories of others. Thus they need to anticipate how others will respond to their decisions. 
The game-theoretic framework provides an appealing model choice to describe this complex decision-making problem among agents~\cite{osborne2004introduction} and has been applied to various robotics applications, e.g., drone racing~\cite{wang2019game} and swarm coordination~\cite{abdelkader2021aerial}.
In an ideal situation, where all the agents are perfectly rational (i.e., they have unlimited computational resources), they can select the motion trajectories that reach the Nash equilibrium (if exists). 
However, since these trajectories live in a continuous space, agents need to evaluate infinitely many trajectories and the interaction among them, which is intractable. 
As a remedy, most of the works consider constraining the search space of the multi-robot problem via sampling~\cite{williams2018best} or locally perturbing the solution~\cite{wang2019game} to find a good trajectory within a reasonable amount of time.

Most game-theoretic planners mentioned above do not explicitly consider the agents' computational limitations in their modeling process, i.e., they assume each agent is perfectly rational.
They consider these limitations only externally, e.g., by truncating the number of iterations during optimization. 
In contrast, we propose a novel and more principled treatment for modeling agents with limited computational resources by directly modeling the agents being only \textit{bounded-rational}~\cite{genewein2015bounded} in the game-theoretic framework.
Bounded Rationality (BR) has been developed in economics~\cite{simon1955behavioral} and cognitive science~\cite{gigerenzer2009homo} to describe behaviors of humans (or other intelligent agents), who have limited computational resources and partial information about the world but still need to make decisions from an enormous number of choices, and has been applied to analyze the robustness of controllers in the single-agent setting~\cite{pacelli2021robust}.
In this work, we use the information-theoretic view of BR~\cite{genewein2015bounded}, which states that the agent optimizes its strategy under an information-theoretic constraint (e.g., KL-divergence), describing the cost of transforming its a-prior strategy into an optimized one. 
This problem can be solved efficiently by evaluating a finite number of trajectory choices from its prior trajectory distribution.
Since BR explicitly considers the computational constraints during the modeling process, the resulting solution provides an explainable way to trade-off computational efficiency and performance.
Furthermore, by incorporating BR into the game-theoretic framework, robots can naturally reason about other agents' sub-optimal behaviors and use these predictions to take advantage of (or avoid) other agents with less (or higher) computational resources.


\section{Related Work}
\label{sec:related_work}
Our work is related to motion planning in multi-agent systems and game-theoretic frameworks.
Here we provide a brief overview of these topics.
The game-theoretic approach in robotics has gained increasing popularity recently. 
For example, the authors in~\cite{wang2019game, spica2020real} combine Model-Predictive Control (MPC) and Iterated Best Response (IBR) to approximate the Nash Equilibrium solution of a two-player vehicle racing game.
Recently, a game-theoretic iterative linear quadratic regulator (iLQR) has been proposed to solve a general-sum stochastic game~\cite{schwarting2021stochastic}. 
In addition, the game-theoretic framework is also used in self-driving vehicles for trajectory prediction~\cite{fisac2019hierarchical, schwarting2018planning} and motion planning among pedestrians~\cite{chen2017socially, lutjens2019safe}.
The above works are all based on the concept of rational agents who are assumed to be able to maximize their utility. 
In contrast, the proposed bounded rational framework explicitly considers the information-processing constraints of intelligent agents. 

Although bounded rational solutions are used  in almost all robotic systems in practice, e.g., anytime planners terminate the computation if the time runs out~\cite{ingrand2017deliberation}, only a few works attempt to model this bounded rational assumption explicitly.
Recently, authors in~\cite{pacelli2021robust} analyze the single-agent robust control performance using the information-theoretic bounded rationality~\cite{ortega2015information, genewein2015bounded}. 
Another closely related but independently developed literature is KL-Control~\cite{botvinick2012planning, kappen2012optimal}.
When computing the optimal policy, it includes an additional information-theoretic cost measured by the KL-divergence between a prior policy distribution and a posterior after optimization. 
This is similar to the effect of the information-theoretic constraints of bounded rationality, where the final optimal control distribution can be sampled from the exponential family using approximate Bayesian inference~\cite{williams2017information, lambert2020stein}.
Although these methods have similar traits, they generally only focus on single-agent problems.
In contrast, our work integrates the bounded rationality idea into the game-theoretic framework and provides a method to compute agents' strategies under computational limits.

\section{Problem Formulation}
\label{sec:problem_formulation}
In this section, we define multi-agent decision-making using the formulation of Multi-Agent Markov Decision Processes (MMDPs) or Markov Game (MGs)~\cite{littman1994markov}, and provide the resulting Nash Equilibrium (NE) solution concept under the perfect rationality assumption. 
In the subsequent sections, we show how this assumption can be mitigated using the BR concept and derive a sampling-based method to find a Nash Equilibrium strategy profile under bounded rationality. 

\subsection{Multi-Agent Markov Decision Process}
In an MMDP with $N$ agents, each agent $i$ has its own state space $s^i \in \mathcal{S}^i$ and action space $a^i \in \mathcal{A}^i$, where $a^i$ and $s^i$ denote the state and action of agent $i$; $\mathcal{S}^i$ and $\mathcal{A}^i$ denote the corresponding spaces.
We denote the joint states and actions of all the agents as $S = [s^1, ..., s^N]$ and $A = [a^1, ..., a^N]$.
The agents progress in the environment as follows. 
At every timestep $t$, all the agents simultaneously execute their actions $A_{t}$ to advance to the next states $S_{t+1}$ according to their stochastic transition function $s^{i}_{t+1} \sim p^i(s^i_{t+1} | S_t, A_t)$.
At the same time, they receive rewards $R_t = [r_t^1, ..., r_t^N]$, where $r_t^i = f^i_t(S_t, A_t)$ is the reward function for agent $i$. 
Each agent's stochastic transition and reward functions depend on all agents' states and actions in the system.
Under the perfectly rational assumption, the goal for agent $i$ is to find a strategy that maximizes an expected utility function
\begin{equation}\label{eq:utility}
\pi_t^{i,*} = \textstyle{\argmax_{\pi_t^i}}U^{i}(S_t, \Pi_t), 
\end{equation}
where 
$U^{i}(S_t, \Pi_t)=\mathbb{E}\Big[\sum_{k=t}^{H+t} r^i_t(S_k, A_k)\Big]$ is agent $i$'s utility function at $t$; $H$ is planning horizon, and $\Pi_t = [\pi_t^1, ..., \pi_t^N]$ denotes the strategy profile for every agent. 
In this work, we assume that the agents' strategies take a specific form: a distribution over the action sequence $\mathbf{a}_t^i \sim \pi_t^i(\mathbf{a}_t^i | S_t, \Pi_t^{-i})$, where $\mathbf{a}_t^{i} = [a_t^i, ..., a_{t+H}^i]$ is the action sequence up to horizon $H$ and $\Pi_t^{-i}$ is the strategy profile without agent $i$. 
This policy form is well-suited for most robotics problems because the trajectory induced by the action sequence can be tracked by a low-level controller. 

\subsection{Iterative Best Response for MMDPs}
To solve the problem defined in Eq.~\eqref{eq:utility}, each agent needs to predict how other agents will behave and respond to each other.
For brevity, we will write $\pi^i(a_t^i | S_t)$ as agent $i$'s strategy and omit the dependency on $\Pi^{-i}$.
One possible and common way to predict other agents' behaviors is by assuming all other agents are perfectly rational, and thus the strategy profile of all the agents reaches the Nash Equilibrium (NE) (if exists)~\cite{osborne2004introduction}, which satisfies the following relationship:
\begin{equation}\label{eq:best-response}
    U^i(S_t, \Pi_t^{-i,*}, \pi_t^{i,*}) \geq 
    U^i(S_t, \Pi_t^{-i,*}, \pi_t^{i}), \forall i \in \{1, ..., N\},
    \text{for any $\pi_t^{i}$,} 
\end{equation}
Intuitively, if the agents satisfy the NE, no agent can improve its utility by unilaterally changing its strategy.
To compute NE, we can apply a numerical procedure called Iterative Best Response (IBR)~\cite{reeves2012computing}. 
Starting from an initial guess of the strategy profiles of all the agents, we update each agent's strategy to the best response to the current strategies of all other agents. 
The above procedure is applied iteratively for each agent until the strategy profile does not change. 
If every agent is perfectly rational, the robot can use NE strategy profile to predict other agents' behaviors and act correspondingly.\footnote{We make a simplifying assumption that there is only one NE in the game.}. 
However, there is a gap between this perfect rational assumption and the real world, as most existing methods can only search the strategy profile in a neighborhood of the initial guess~\cite{wang2019game, spica2020real} due to computational limits.
In the following section, we fill this gap by explicitly formulating this bounded rational solution.

\section{Methodology}
\label{sec:methodology}
This section first provides details on the formulation of bounded rationality and integrates it into the game-theoretic framework.
Then, we propose a sampling-based approach to generate bounded-rational stochastic policies. 

\subsection{Bounded Rational Agents in Game-Theoretic Framework}
In the standard game-theoretic frameworks, agents are rational, i.e., they optimize their decisions via evaluating an infinite number of action sequences without considering the computational resources. 
In contrast, we model each agent as bounded rational -- it makes decisions that maximize its utility subject to a certain computational constraint. 
Following the work in information-theoretic bounded rationality~\cite{ortega2013thermodynamics, genewein2015bounded}, this constraint is explicitly defined by the neighborhood of a default policy $q^i$.
Intuitively, this default policy describes the nominal behavior of the agent. 
For example, in a driving scenario, the default policy of an aggressive driver may be more likely to drive at a high speed. 
The agent's goal is to search for an optimized posterior policy bounded within the neighborhood of $q^i$.  
This size of the neighborhood may reflect the limited computational resources or other practical considerations. 
In the following, we omit the time subscript $t$ for clarity. 
We use KL-divergence to characterize this neighborhood 
\begin{equation}\label{eq:constrained-value-fn}
\begin{split}
    \pi^{i, *} = \argmax_{\pi^{i}} U^i(S, \Pi) 
    \textrm{, s. t. } KL(\pi^{i} || q^{i}) \leq K_i.
\end{split}
\end{equation}
 $K_i$ is a constant denoting the amount of computation (measured in bits) agent $i$ can deviate from the default policy. 
Using Lagrange multipliers, we can rewrite the constrained optimization problem in Eq.~\eqref{eq:constrained-value-fn} as an unconstrained one
$\pi^{i, *} = \argmax_{\pi^i} U^i(S, \Pi) - \frac{1}{\beta_i} KL(\pi^{i} || q^{i})$, 
where $\beta_i > 0$ indicates the {\em rationality level}.
To see how this bounded-optimal stochastic policy can be computed, we can first observe that the unconstrained problem can be written as  
\begin{equation}\label{eq:proof-distribution}
\begin{split}
  &U^i(S, \Pi) - \frac{1}{\beta_i} KL(\pi^{i} || q^{i}) \\ 
    &= -\frac{1}{\beta_i}
    \Big(KL(\pi^i || q) - \beta U^i(S, \Pi) \Big) \\
    &= -\frac{1}{\beta}
    \int \pi^{i}(\mathbf{a}^i | S)  
            \Big(\log \frac{\pi^{i}(\mathbf{a}^i | S) }{q^{i}(\mathbf{a}^i)e^{\beta U^i(S, \Pi)}} \Big)d\mathbf{a}^{i} 
    \\
    &= -\frac{1}{\beta}KL(\pi^i || \psi^i),
\end{split}
\end{equation}
where $\psi^i(\mathbf{a}^i | \mathbf{s}) \propto q^{i}(\mathbf{a}^i)e^{\beta U(\mathbf{s}, \mathbf{a})}$.
Since KL-divergence is non-negative, the maximum of $-\frac{1}{\beta}KL(\pi^i||\psi^i)$ is obtained only when $KL(\pi^i || \psi^i) = 0$, which means $\pi^i = \psi^i$.
Therefore, the optimal action sequence distribution of agent $i$ (while keeping other agents' strategies fixed) under the bounded rationality constraint is 
\begin{equation}\label{eq:optimal}
    \pi^{i, *}(\mathbf{a}^i | S, \Pi^{-i}) = \frac{1}{Z} 
    q^i(\mathbf{a}^i)e^{\beta\cdot U^i(S, \Pi)},
\end{equation}
where $Z = \int q^i(\mathbf{a}^i)e^{\beta\cdot U^i(S, \Pi)} d\mathbf{a}^i$ is a normalization constant.
This bounded-optimal strategy provides an intuitive and explainable way to trade-off between computation and performance. 
When $\beta$ increases from $0$ to infinity, the agent becomes more rational and requires more time to compute the optimal behavior.
When $\beta_i = 0$, the bounded-rational policy becomes the prior, meaning agent $i$ has no computational resources to leverage. 
On the other hand, when $\beta_i \rightarrow \infty$, the agent becomes entirely rational and selects the optimal action sequence deterministically.  
The rationality parameter $\beta$ allows us to model agents with different amounts of computational resources.

Similar to the rational case, our goal is to find the Nash Equilibrium strategy profile for a group of bounded-rational agents whose bounded-optimal policies are defined in Eq.~\eqref{eq:optimal}. 
This can be done using the IBR procedure analogous to Section~\ref{sec:problem_formulation}.
Instead of optimizing the policy in Eq.~\eqref{eq:utility}, each bounded-rational agent finds the optimal strategy distribution defined in Eq.~\eqref{eq:optimal} while keeping other agents' strategies fixed.
This procedure is carried out for each agent for a fixed number of iterations or until no agent's stochastic policy changes. 


\subsection{Importance Sampling for Computing Bounded-Rational Strategies}
The previous section describes the bounded rationality concept, its integration with the game-theoretic framework, and uses the IBR numerical method to solve for a bounded rational Nash Equilibrium strategy profile.
To actually compute the bounded-rational strategies, we need an efficient way to query samples from the distribution in Eq.~\eqref{eq:optimal} for each agent.
Since it is relatively easier to sample the actions from the default $q^i(\mathbf{a}^i)$, we can use importance sampling to estimate the expectation of the optimal action sequence as the best response for the agent $i$ while keeping others' actions fixed~\cite{bishop2006pattern}
\begin{equation}\label{eq:expectation-action}
\begin{split}
\mathbb{E}_{\mathbf{a}^{i,*}_{t} \sim \pi^{i,*}_{t}}[\mathbf{a}^{i, *}_{t} | S_t, \Pi^{-i}_t] &= \frac{1}{Z} \int 
  \mathbf{a}^i_t q^i_{t}(\mathbf{a}^i_t) e^{\beta U^i(S_t, \Pi_t)} d\mathbf{a}^i_t \\
   &= \frac{1}{Z}\mathbb{E}_{\mathbf{a}^{i}_t \sim q^i_{t}(\mathbf{a^i_t})}[w(\mathbf{a}^i_{t})\mathbf{a}^i_t] \\
   &\approx \frac{1}{Z}\frac{1}{K} \sum_{k=1}^{K} w(\mathbf{a}^i_{t, k})\mathbf{a}^i_{t, k},
\end{split}
\end{equation}
where $w(\mathbf{a}^i_t) = \exp\{\beta \sum_{k=t}^{H+t} r^i_t(S_k, A_k)\}$ and $\mathbf{a}^i_{t,k}$ denotes the $k^{th}$ sample from the default policy with $N$ samples in total. 
Similarly, the normalization constant can also be approximated as 
\begin{equation}\label{eq:normalization-const}
    \begin{split}
        Z &= \int q^i(\mathbf{a}_t^i)e^{\beta U^i(S_t, \Pi_t)} d\mathbf{a}^i_t \\
          &= \mathbb{E}_{\mathbf{a}^i_t \sim q^i(\mathbf{a}^i_t)}[w(\mathbf{a}^i_t)] \\ 
          & \approx \frac{1}{K} \sum_{k=1}^{K} w(\mathbf{a}^i_{t, k}). 
    \end{split}
\end{equation}
At a high level, the importance sampling procedure proceeds as follows.
The agents first propose action sequence samples from their default policies $q^i_t(\mathbf{a}^i)$ and then assign each sequence a weight $w(\mathbf{a_t}^i)$ indicating its value based on the agents' utilities and rationality levels. 
Finally, by combining Eq.~\eqref{eq:expectation-action} and Eq.~\eqref{eq:normalization-const}, we can use the weighted average to compute the expected optimal action sequence
\begin{equation}\label{eq:importance-sampling}
\begin{split}
    \mathbb{E}_{\mathbf{a}^{i,*}_{t} \sim \pi^{i,*}_{t}}[\mathbf{a}^{i, *}_{t} | S_t, \Pi^{-i}_t]  
           \approx \frac{\sum_{k=1}^{N} w(\mathbf{a}^{i}_{t,k}) \mathbf{a}^i_{t,k}}{\sum_{k=1}^{N} w(\mathbf{a}^{i}_{t,k})}.
\end{split}
\end{equation}
To find the bounded-rational Nash Equilibrium strategy profile, we replace the optimization procedure in IBR Eq.~\eqref{eq:best-response} using the above importance sampling.
One important observation is that the shape of the prior distribution $q^i$, the number of samples for evaluation $N$, and the rationality level $\beta$ play essential roles in the final performance.
Their relationships will be explored in the experimental section. 

\section{Simulated Experiments}
\label{sec:experiment}

We conduct extensive simulation experiments to demonstrate that integrating bounded rationality with the game theory framework (1) allows each agent to reason about other agents' rationality levels to exploit (or avoid) others with less (or higher) computational capabilities and (2) explicitly trades-off between the performance and computation by varying the rationality level $\beta$ and the number of sampled trajectories.
We also show qualitatively that our method can generate group behaviors that approximately reach a bounded rational Nash Equilibrium strategy profile for a varying number of agents.

\subsection{Simulation Setup}
In this experiment, we consider the task of navigating a group of aerial vehicles in a 3D space. 
Each agent’s goal is to swap its position with the diametrically opposite one while avoiding collisions with each other.
The distance between each agent and its goal is $6m$. 
This environment is neither fully cooperative as each agent has a different goal location nor fully competitive because their objectives are not totally reversed (zero-sum).
Thus, the agents need to collaborate with each other to avoid blockage in the center and at the same time compete with each other to follow their shortest paths to the goals.
The agents are homogeneous in terms of their sizes and physical capabilities.
Each agent has a size of $0.125m$ in $x,y,z$ directions (similar to the dimension of Crazyfly drones used in the next section).
Similar to~\cite{wang2019game}, we set their transition functions using a deterministic single integrator model with the minimum and maximum speeds as $a_{min} = 0 m/s$ and $a_{max} = 1m/s$. 
We set a uniform prior $q^i(\mathbf{a}) = Uniform(a_{min}, a_{max})$ as the default policy for all the agents. 
The number of IBR iterations is set to $10$ as we found that under varying parameters, IBR usually converges to a bounded rational NE strategy at $10$ iterations.
In all the simulations, we assume that the rationality levels of all the agents are known to each other.
The one-step reward function is the same for each agent and set to penalize collisions and large distances to the goal.
We run $50$ times for each simulation with $T=80$ timesteps. 

\begin{figure*}[t] \vspace{-10pt}
  \centering
  \subfigure[]
  	{\includegraphics[height=1.2in, width=1.5in]{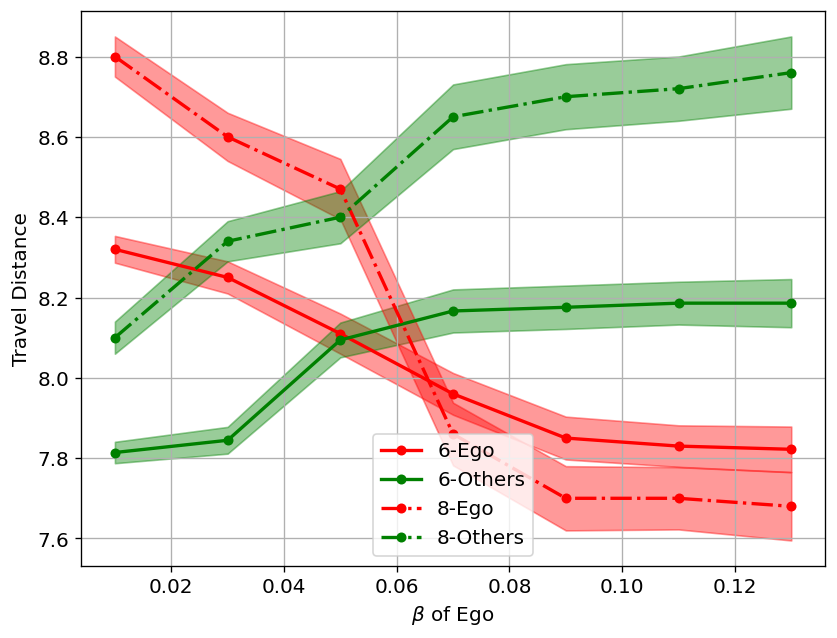}
         \label{fig:sim_6cf_betas}}
  \subfigure[Ego's $\beta=0.01$]
  	{\includegraphics[height=1.2in, width=1.5in]{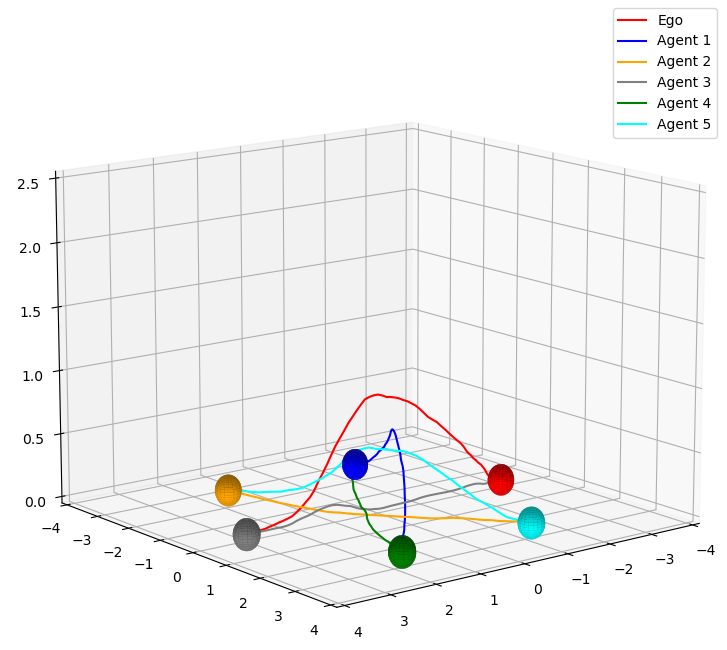}
         \label{fig:sim_6cf_01}}
  \subfigure[Ego's $\beta=0.13$]
  	{\includegraphics[height=1.2in, width=1.5in]{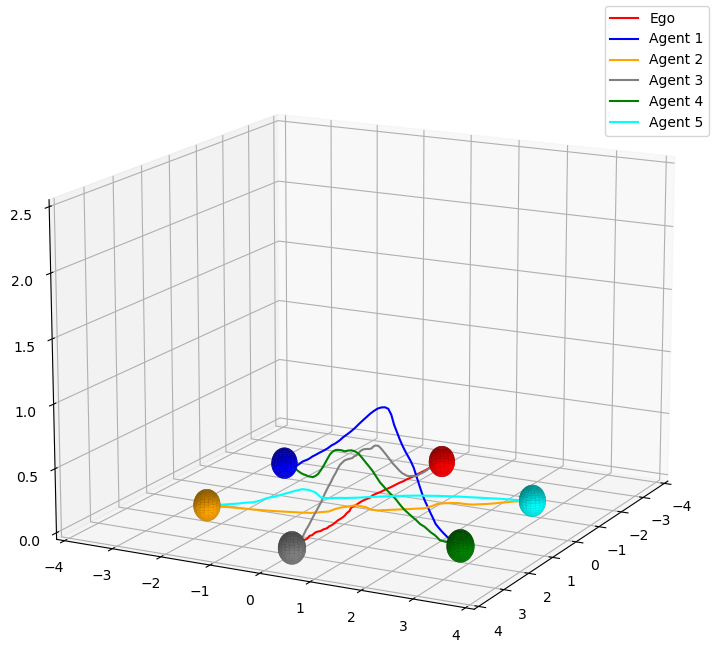}
         \label{fig:sim_6cf_13}}
  \caption{(a) Comparison of the performance by increasing the ego's rationality level from $\beta=0.01$ to $\beta=0.13$ while keeping other agents' rationality levels fixed in six (solid lines) and eight (dashed lines) agent environments. 
  The x-axis and y-axis indicate the ego's $\beta$ values and traveled distances.
  The green lines are the average travel distance of other agents and the red lines indicate the ego's travel distance.
  (b)(c) Show the agents' trajectories in six-agent environment with $\beta=0.01$ and $\beta=0.13$, respectively.
  The ego's trajectories are depicted in red.
  \vspace{-10pt}}
\label{fig:sim_6cf_swap}  
\end{figure*}

\subsection{Results}
We first show that using the proposed framework agents can naturally reason about the behaviors of others with the same physical capabilities but different rationality levels $\beta$.
Since we want to examine how varying $\beta$ affects the performance, we need to ensure that the policy converges to the ``optimal" one under the computational constraints.
Thus, we sample a large number of trajectories, $5\times 10^5$, for policy computation for each agent. 
In this simulation, we fix other agents' $\beta = 0.05$ and vary one agent's (called {\em ego} agent) $\beta$ from $0.01$ to $0.13$ and compare the travel distances between the robot and other agents (the distance of other agents is averaged).
Fig.~\ref{fig:sim_6cf_swap}\subref{fig:sim_6cf_betas} shows the performance comparison in six and eight agent environments.
In general, when the ego has a lower rationality level $\beta$ than other agents, it avoids them by taking a longer path. 
When all the agents have the same $\beta$, the ego and other agents have a similar performance. 
As $\beta$ increases, the ego begins to exploit its advantage in computing more rational decisions and taking shorter paths.
We also notice that the ego generally performs better with a large $\beta$ when there are more agents in the scene. 
The trajectories of the six-agent environment for $\beta=0.01$ and $\beta=0.13$ are plotted in Fig~\ref{fig:sim_6cf_swap}\subref{fig:sim_6cf_01} and  Fig~\ref{fig:sim_6cf_swap}\subref{fig:sim_6cf_13}, respectively.
When the ego's $\beta=0.01$ (in the red trajectory), it takes a detour by elevating to a higher altitude to avoid other agents.
In contrast, when its $\beta=0.13$, it pushes other agents aside and takes an almost straight path.
These results are aligned with Fig.~\ref{fig:sim_6cf_swap}\subref{fig:sim_6cf_betas}. 
We omit the trajectories of eight agent environments to avoid clutter.
The readers are encouraged to watch the videos at \url{https://youtu.be/hzCitSSuWiI}

\begin{figure*}[t!] 
  \centering
  \subfigure[]
  	{\includegraphics[height=1.5in, width=2.1in]{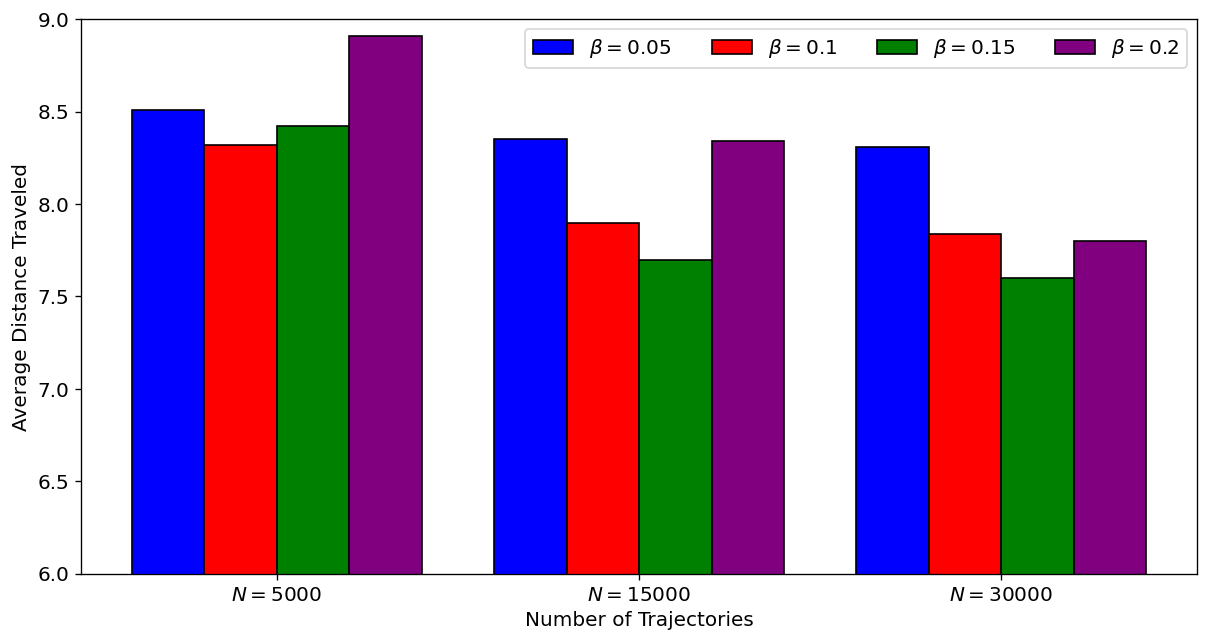}
         \label{fig:tradeoff-6}}
  \subfigure[]
  	{\includegraphics[height=1.5in, width=2.1in]{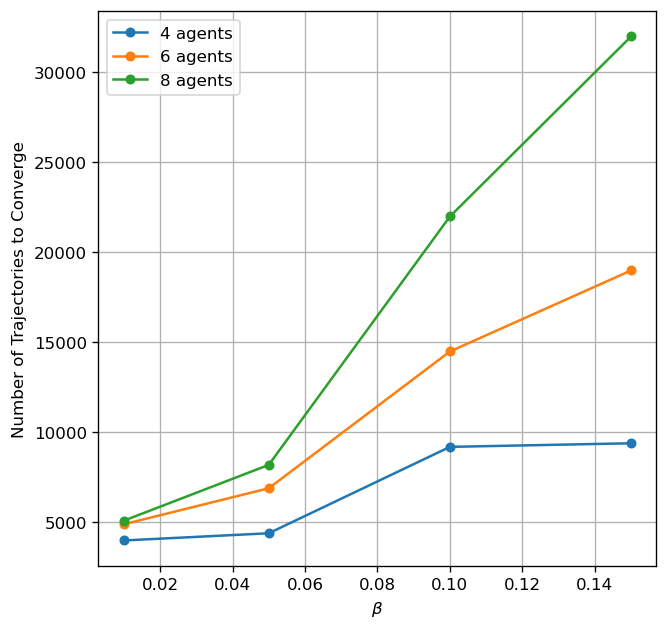}
         \label{fig:convergence}}
  \caption{ (a) Compares the performance (average traveled distance of the group) of different $\beta$ values using a different number of trajectories in the six-agent environment. 
  E
  The $x$ and $y$ axes are the number of trajectories and the average traveled distance of the group. 
  (b) Shows the number of trajectories required to converge for different $\beta$ in four, six, and eight agent environments.
  \vspace{-10pt}}
\label{fig:computational_tradeoff}  
\end{figure*}

\begin{figure*}[t!] 
  \centering
  \subfigure[]
  	{\includegraphics[height=1.2in, width=1.5in]{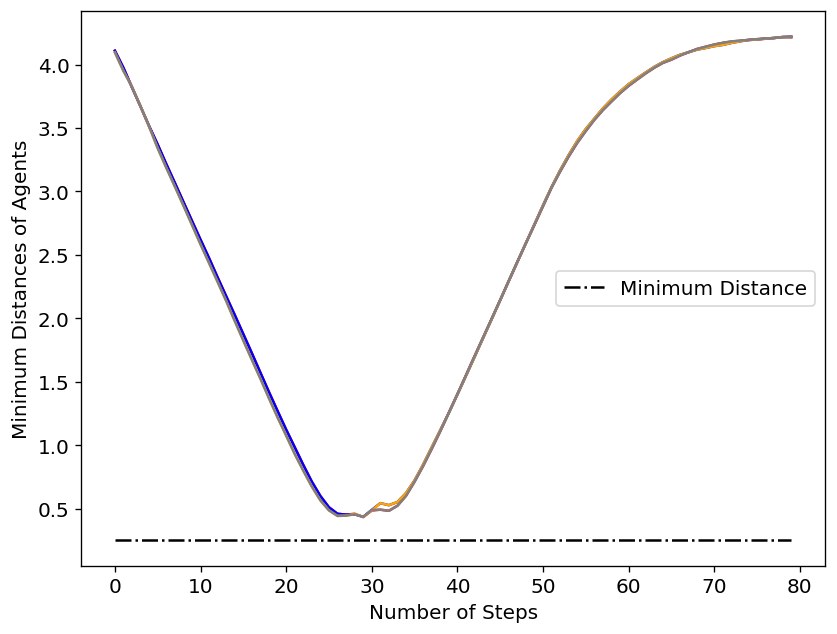}
         \label{fig:tradeoff-6}}
  \subfigure[]
  	{\includegraphics[height=1.2in, width=1.5in]{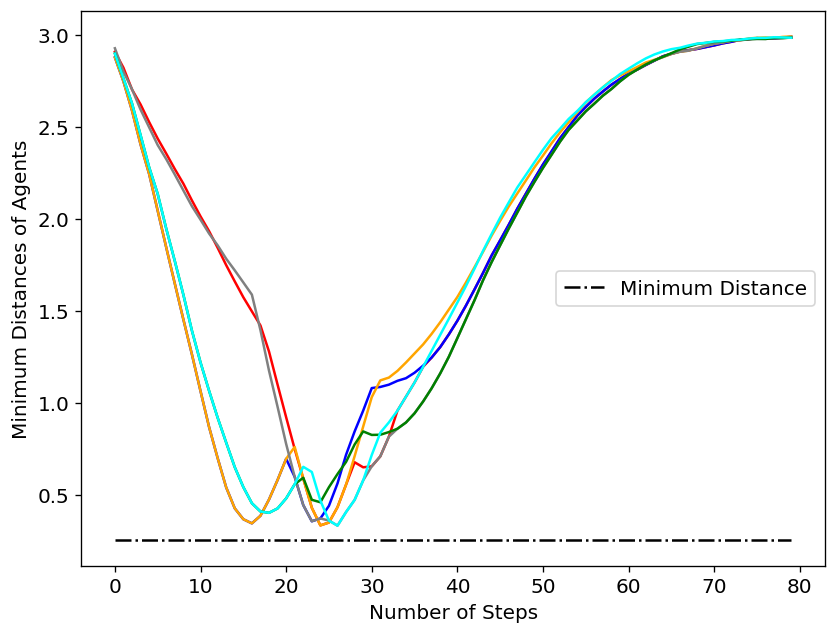}
         \label{fig:tradeoff-6}}
  \subfigure[]
  	{\includegraphics[height=1.2in, width=1.5in]{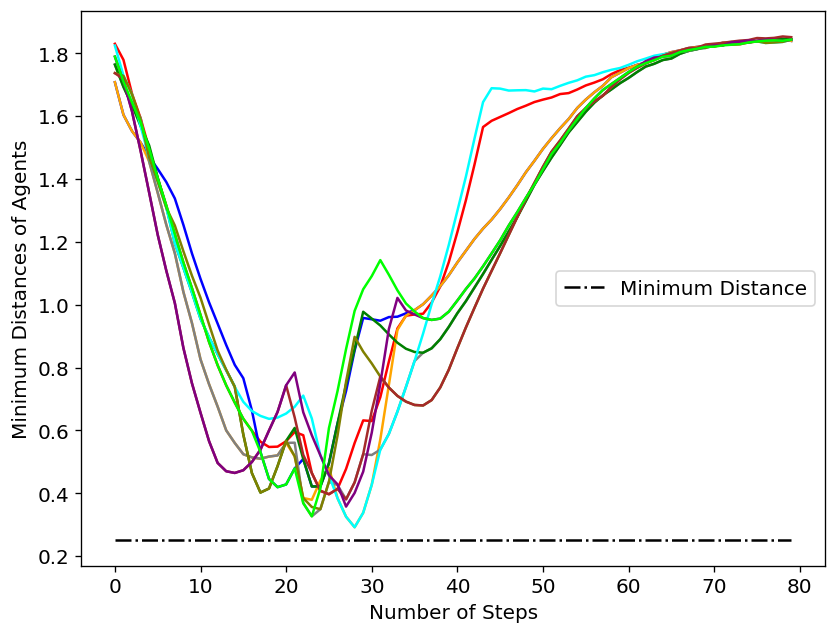}
         \label{fig:tradeoff-6}}
  \caption{Minimum distances of each agent to other agents at every timestep in (a) four-agent (b) six-agent (c) ten-agent environments.
  The x-axis and y-axis are the timesteps and agents' minimum distances, respectively. 
  The colored solid lines represent the statistics of each agent, which is the same as their trajectory color in Fig.~\ref{fig:4-agent-traj}. 
  The dashed grey line shows the minimum safety distance ($0.25m$) that needs to be maintained to avoid collisions. 
  \vspace{-10pt}}
\label{fig:agents_distances}  
\end{figure*}

\begin{figure*}[h!] 
  \centering
  \makebox[20pt]{\raisebox{40pt}{\rotatebox[origin=c]{90}{Four agents}}}
  \subfigure
  	{\includegraphics[height=1.2in, width=1.4in]{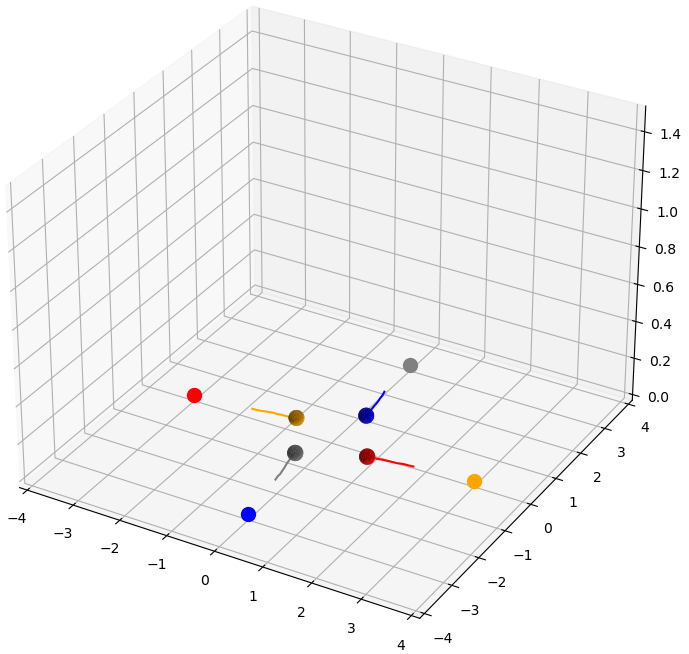}
         \label{fig:tradeoff-6}
         }
  \subfigure
  	{\includegraphics[height=1.2in, width=1.4in]{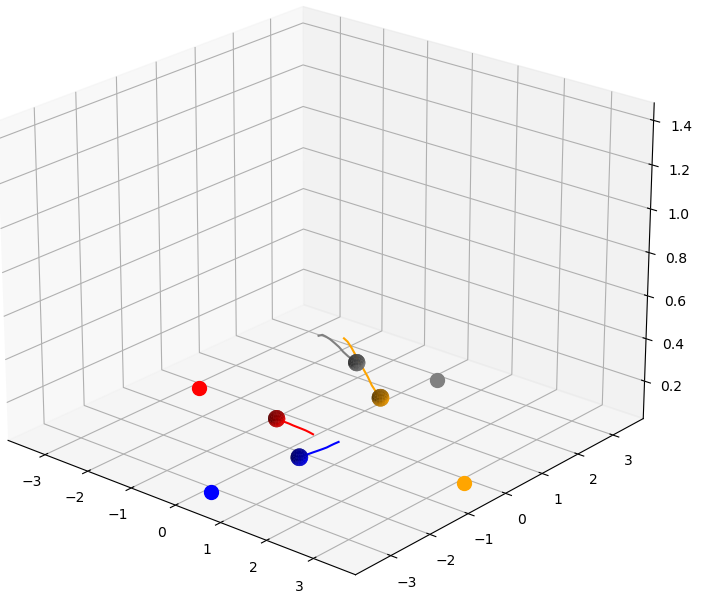}
         \label{fig:tradeoff-6}}
  \subfigure
  	{\includegraphics[height=1.2in, width=1.4in]{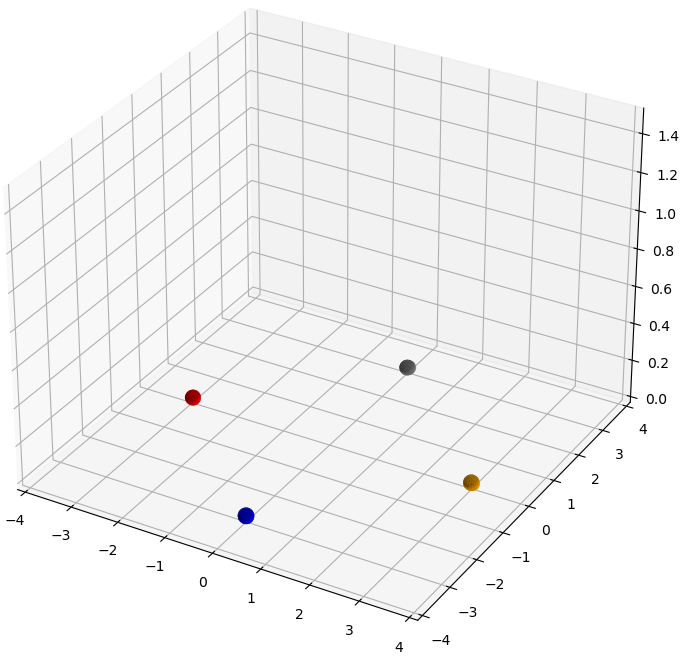}
         \label{fig:tradeoff-6}}\\
  \makebox[20pt]{\raisebox{40pt}{\rotatebox[origin=c]{90}{Six agents}}}
  \subfigure
  	{\includegraphics[height=1.2in, width=1.4in]{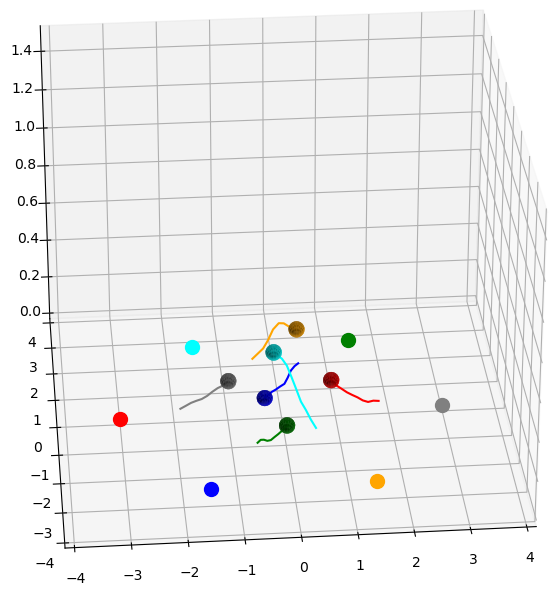}
         \label{fig:}}
  \subfigure
  	{\includegraphics[height=1.2in, width=1.4in]{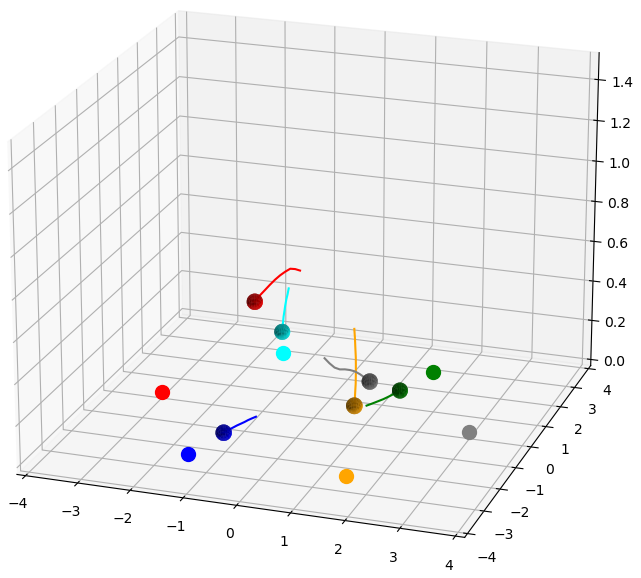}
         \label{fig:}}
  \subfigure
  	{\includegraphics[height=1.2in, width=1.4in]{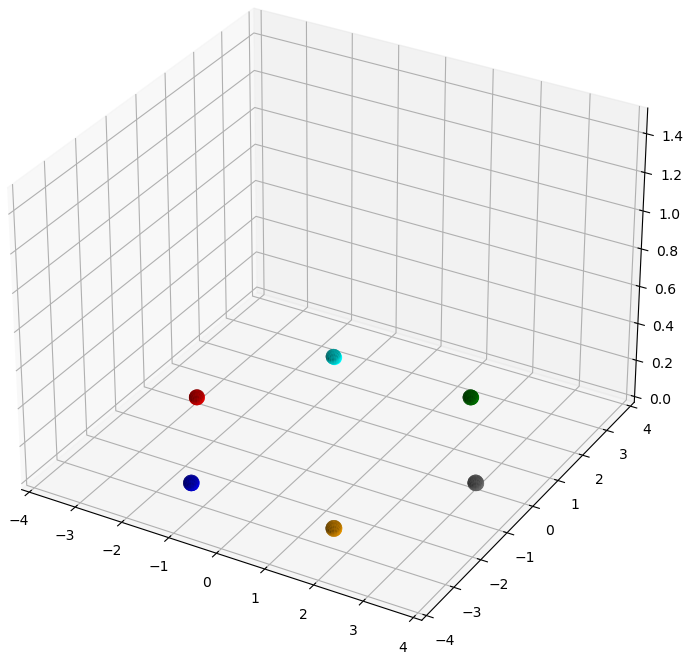}
         \label{fig:}}\\
  \makebox[20pt]{\raisebox{40pt}{\rotatebox[origin=c]{90}{Ten agents}}}
  \subfigure
  	{\includegraphics[height=1.2in, width=1.4in]{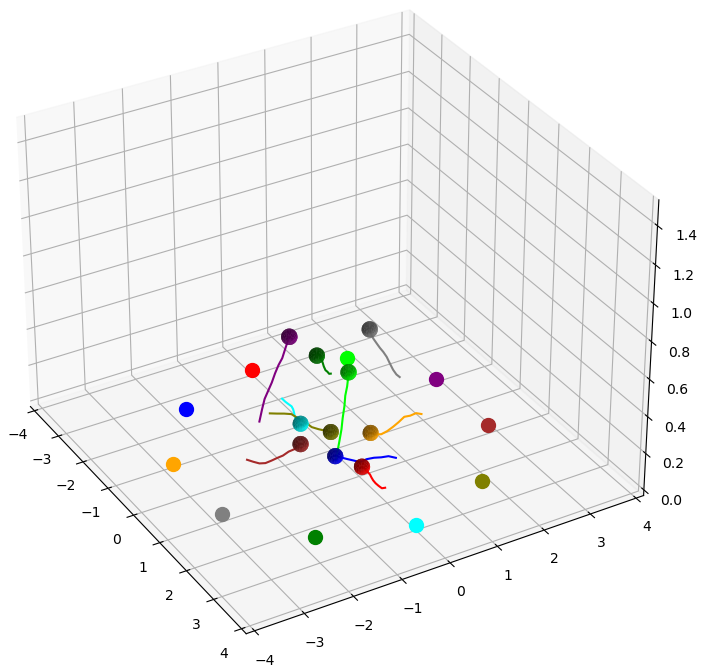}
         }
  \subfigure
  	{\includegraphics[height=1.2in, width=1.4in]{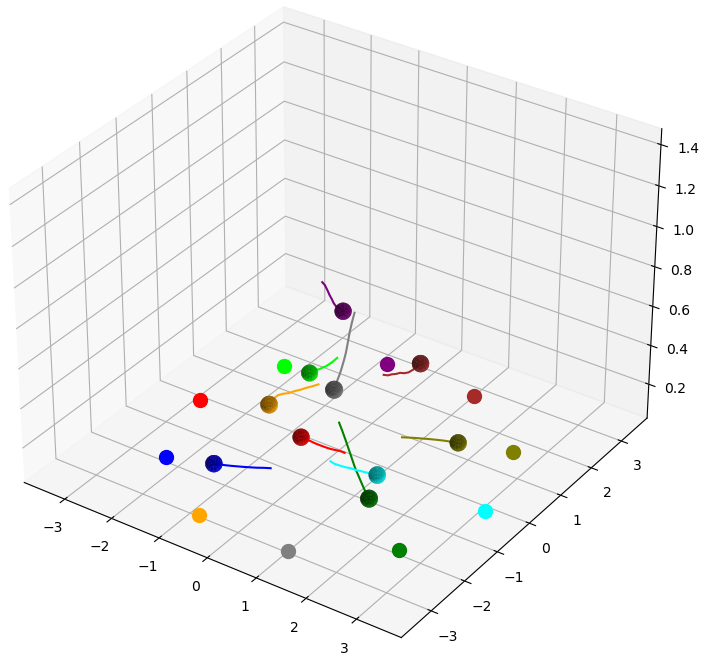}
         \label{fig:}}
  \subfigure
  	{\includegraphics[height=1.2in, width=1.4in]{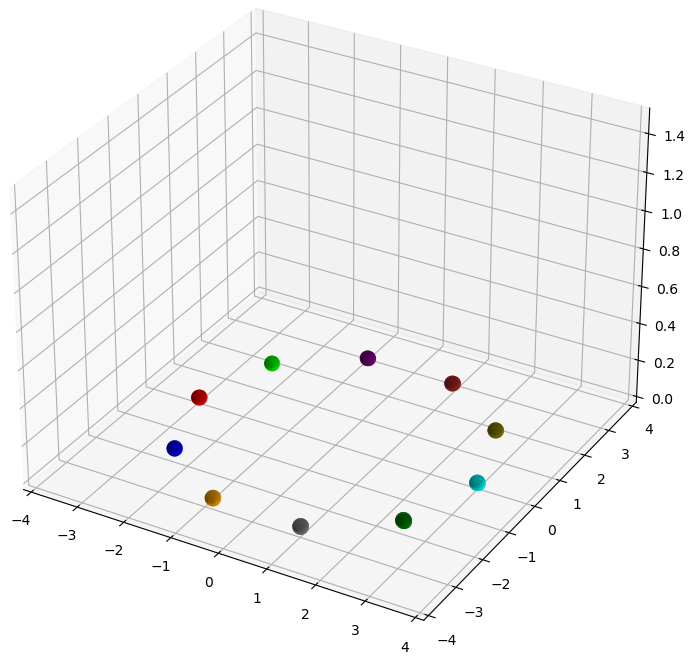}
         \label{fig:}}
  \caption{ 
  Agents' trajectories in environments with four, six, and ten agents. 
  The three columns show the snapshots at $t=20, 60, 80$, respectively.
  Each agent, its trajectory, and its goal are assigned the same unique color.
  We only show the last $10$ steps of trajectories to avoid clutter.
  \vspace{-10pt}}
\label{fig:4-agent-traj}  
\end{figure*}

Next, we evaluate the performance of a group of agents with the same $\beta$ to show that the trade-off between the performance and computation can be directly controlled by the rationality level.
Since most of the computation occurs when evaluating a large number of sampled trajectories in the proposed importance sampling-based method, we can use the number of evaluated trajectories as a proxy to measure the amount of computation an agent possesses. 
In Fig.~\ref{fig:computational_tradeoff}\subref{fig:tradeoff-6}, we analyze the relationship between $\beta$ and the computation constraint in the six-agent environment. 
We can observe that when the computation is limited, i.e., only a small number of action sequences can be evaluated, a larger $\beta$ (more rational) actually hurts the performance.
When the more rational agents have the resources to evaluate more trajectories, they travel less distance on average than the less rational groups.  
This result demonstrates that by controlling the rationality parameter the bounded rational framework can effectively trade-off between optimality and limited computation. 
In Fig.~\ref{fig:computational_tradeoff}\subref{fig:convergence}, we also evaluate the number of trajectories that need to be sampled for the method to converge at different rationality levels in four, six, and eight agent environments.
The result aligns with the previous observation -- in general, when $\beta$ is larger, more trajectories need to be evaluated to converge to ``optimal" under the bounded rational constraints. 
Furthermore, when more agents are present in the environment, the method requires more trajectories to converge.

Finally, we show qualitative trajectory results of the group behaviors under bounded rationality using a fixed $\beta=0.1$ and number of samples $N=5\times 10^5$ for each agent in Fig.~\ref{fig:4-agent-traj}.
Additionally, we provide the minimum distances of each agent to other agents in Fig.~\ref{fig:agents_distances}. 
The result shows that in each environment, the agent can maintain a safe distance $>0.25m$ to avoid collisions.

\section{Physical Experiments}

\begin{figure}[t!] 
  \centering
  \subfigure[]
  	{\includegraphics[height=1.5in, width=2.33in]{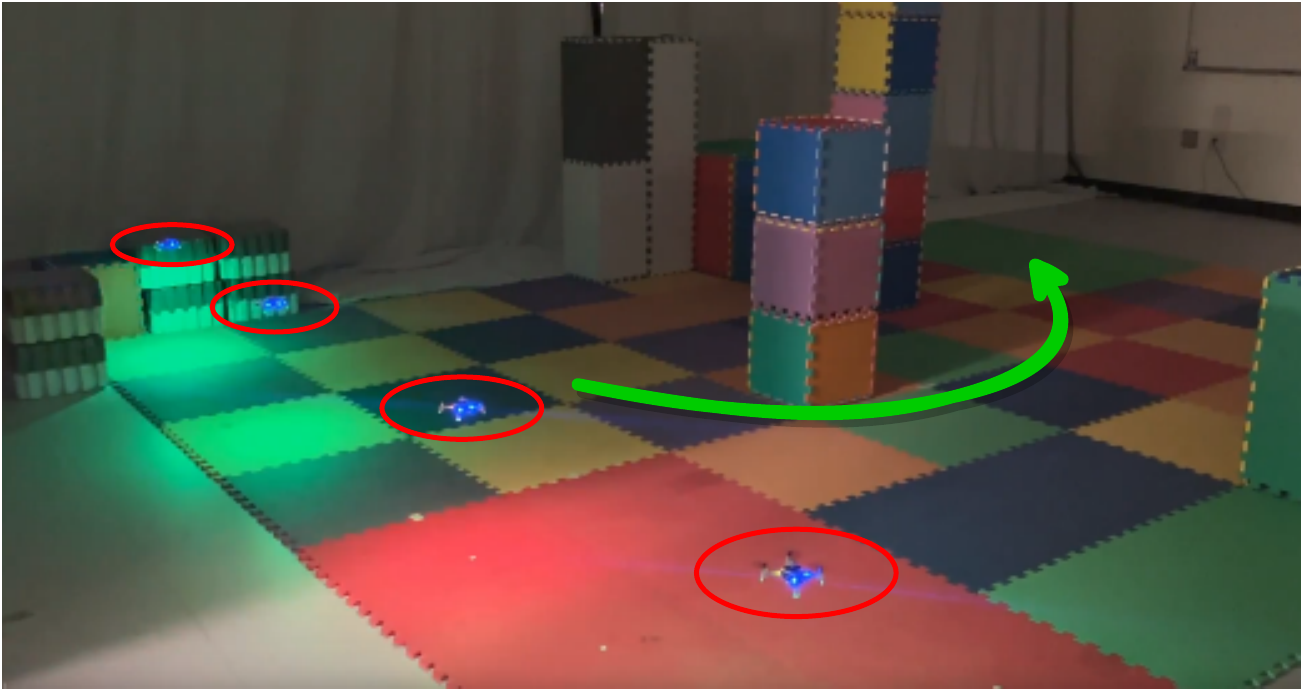}
         \label{fig:4cf_setup}}
   \subfigure[]
  	{\includegraphics[height=1.5in, width=2.33in]{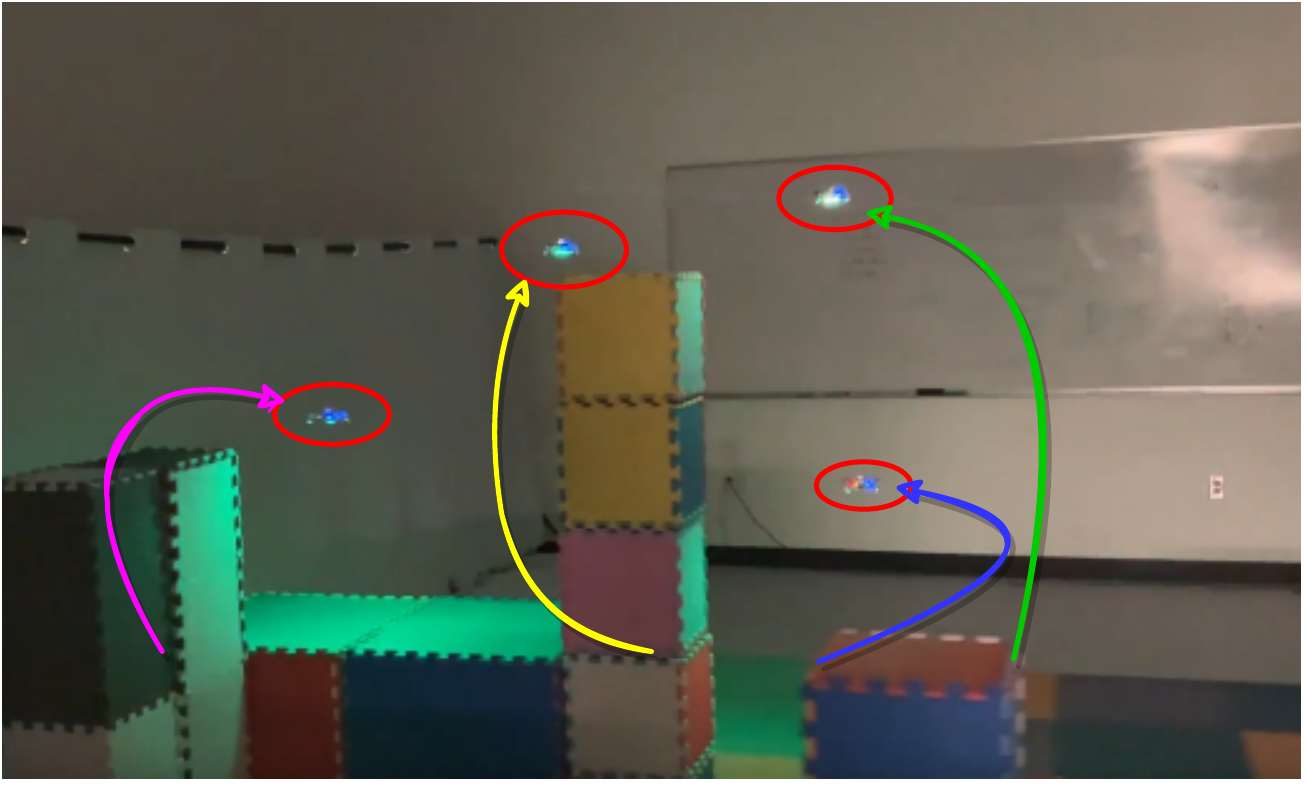}
         \label{fig:4cf_exp}}
  	\vspace{-10pt}
  \caption{(a) Shows the experimental setup with 4 Crazyflie drones. Green arrow points to the goal position. All the agents are navigating to the same goal point. (b) Snapshot of the experiment. Shows the path followed by the agents to avoid the obstacles.   
  \vspace{-10pt}}
\label{fig:4cf_obs_avoid}  
\end{figure}

\begin{figure}[t] 
  \centering
  \subfigure[]
  	{\includegraphics[height=1.5in, width=2.33in]{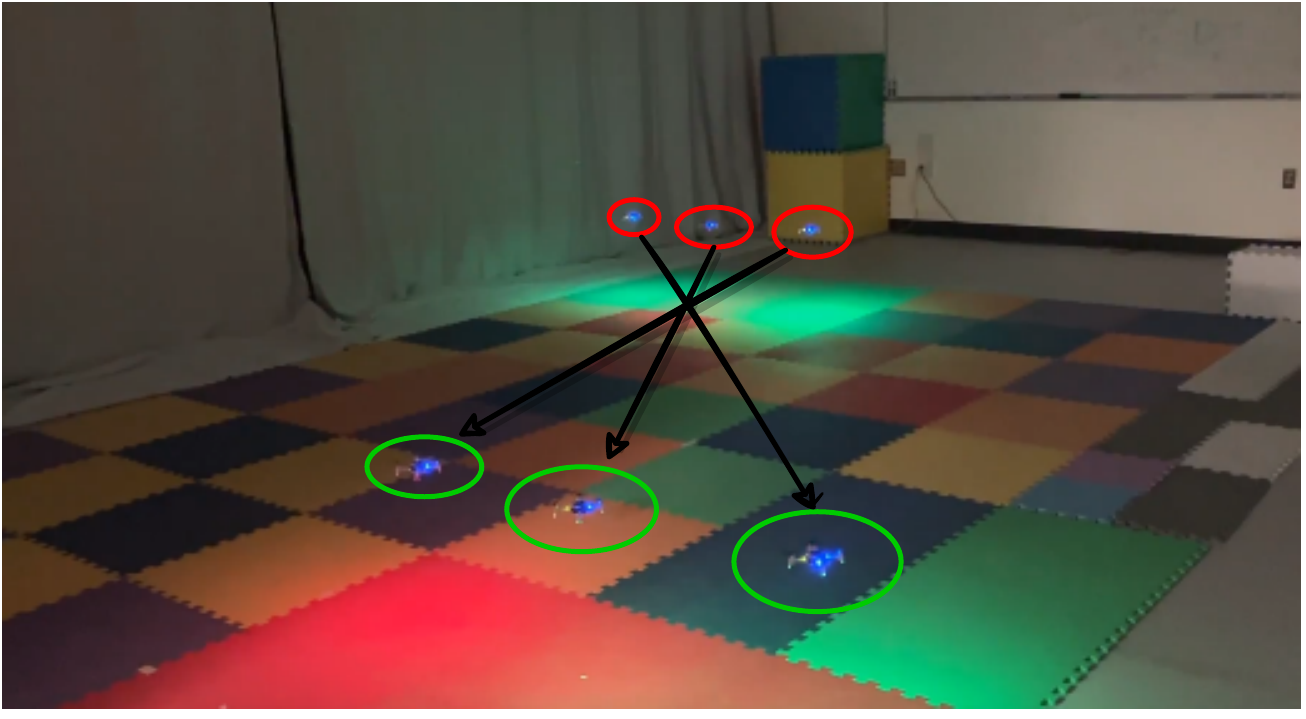}
         \label{fig:6cf_setup}}
   \subfigure[]
  	{\includegraphics[height=1.5in, width=2.33in]{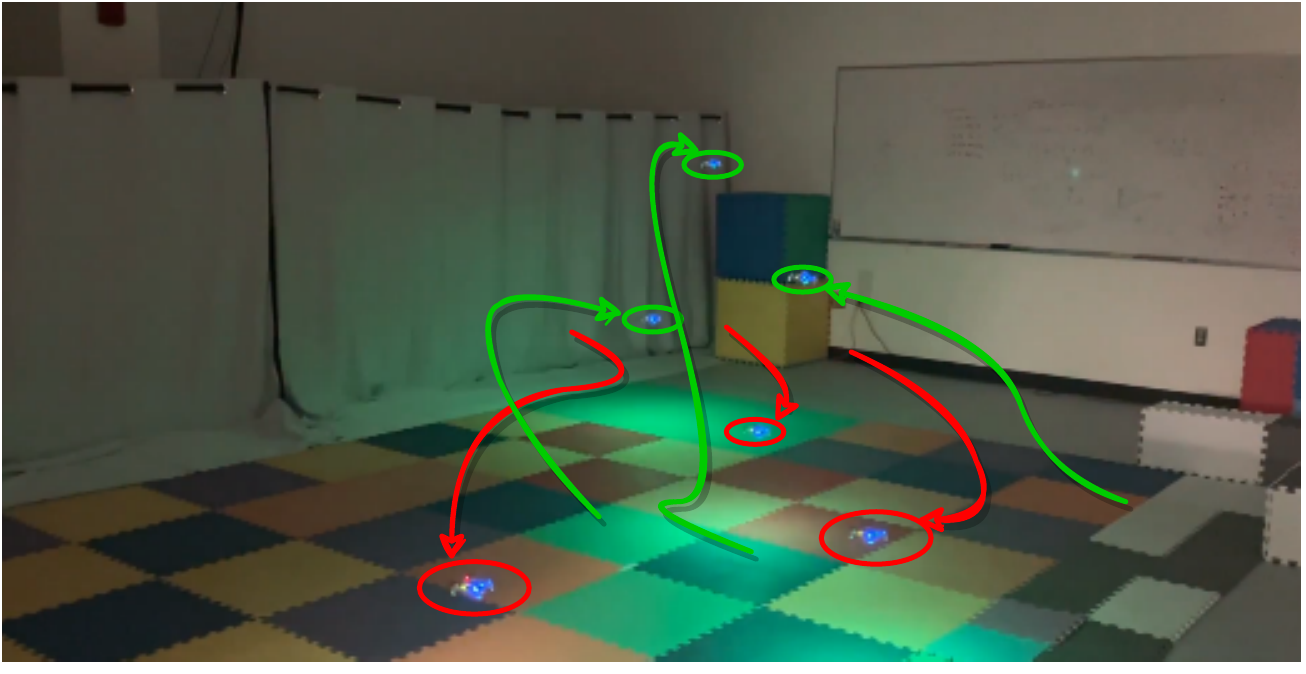}
         \label{fig:6cf_exp}}
  	\vspace{-10pt}
  \caption{Shows the experimental setup with 6 Crazyflie drones. The drones are divided into two groups - red and green. (a) Shows the initial position of the drones. The task is to swap the positions. Black lines show assigned swapping tasks among agents. (b) Snapshot of the experiment during swapping. It shows the path followed by the agents to avoid collision with each other.   
  \vspace{-10pt}}
\label{fig:6cf_position_swap}  
\end{figure}

We use the Crazyflie 2.1 
nano-drones under a motion capture system 
to validate our method in the real world. 
For this hardware experiment, we consider two types of tasks with a varying number of agents.
The first task is to navigate a group of drones to a designated goal region while avoiding static obstacles and inter-drone collisions. 
The second task is position swapping similar to the previous section. 
The size of the workspace considered in all the experiments is $4.2m \times 5.4m \times 2m$ in the x, y, and z axes, and the size of the drone is $92mm \times 92mm \times 29mm$.
To mitigate the downwash effect and control inaccuracy, we buffer the drone's collision size to be $0.5m\times 0.5m\times 0.5m$.
We use Crazyswarm \cite{crazyswarm} platform to control the drones. 
For each run, we generate trajectories using $N=3\times 10^5$ and $\beta = 0.1$ for all the agents using the proposed method.
These trajectories contain a sequence of $(x,y,z)$ coordinates. 
We use minimum snap trajectory optimization and control strategy \cite{mellinger2011} to track the trajectories generated by the proposed planner.    

We show two representative scenarios for each task. 
For complete experimental videos, please refer to \url{https://youtu.be/hzCitSSuWiI}
Fig. \ref{fig:4cf_obs_avoid} shows that a group of four drones have to go from the red zone to the green zone while avoiding four obstacles of various sizes distributed around the center of the workspace. 
Note that one of the drones opted to go over the obstacle of height 1.5m which shows that it finds a path through the space as narrow as the size of the drone ($0.5m$) in the $z$-axis. This event is captured in the snapshot shown in Fig.~\ref{fig:4cf_exp}. Fig.~\ref{fig:6cf_position_swap} shows the position swapping scenario. We use the same dynamics model as the previous section to generate the trajectories. We observe that the outcomes of the physical experiments are consistent with the results obtained in the simulation.

\section{Conclusion} 
\label{sec:conclusion}


This paper considers the problem of making sequential decisions for agents with finite computational resources, where they need to interact with each other to complete their designated tasks.
This problem is challenging because each agent needs to evaluate its infinite number of decisions (e.g., waypoints or actuator commands) and reason how others will respond to its behavior. 
While the game-theoretic formulation provides an elegant way to describe this problem, it is based on an unrealistic assumption that agents are perfectly rational and have the ability to evaluate the large decision space. 
We propose a formulation that replaces this rational assumption with the bounded rationality concept and presents a sampling-based approach to computing agents' policies under their computational constraints.
As shown in the experiments, by removing the perfect rational assumption, the proposed formulation allows the agents to take advantage of those with less computational power or avoid those who are more computational-capable.
Additionally, when all the agents are similarly computational capable, they exhibit behaviors that avoid being taken advantage of by others.

{
\small
\bibliographystyle{plain}
\bibliography{references}

\begin{thebibliography}{10}

\bibitem{abdelkader2021aerial}
Mohamed Abdelkader, Samet G{\"u}ler, Hassan Jaleel, and Jeff~S Shamma.
\newblock Aerial swarms: Recent applications and challenges.
\newblock {\em Current Robotics Reports}, 2(3):309--320, 2021.

\bibitem{bishop2006pattern}
Christopher~M Bishop and Nasser~M Nasrabadi.
\newblock {\em Pattern recognition and machine learning}, volume~4.
\newblock Springer, 2006.

\bibitem{botvinick2012planning}
Matthew Botvinick and Marc Toussaint.
\newblock Planning as inference.
\newblock {\em Trends in cognitive sciences}, 16(10):485--488, 2012.

\bibitem{chen2017socially}
Yu~Fan Chen, Michael Everett, Miao Liu, and Jonathan~P How.
\newblock Socially aware motion planning with deep reinforcement learning.
\newblock In {\em 2017 IEEE/RSJ International Conference on Intelligent Robots
  and Systems (IROS)}, pages 1343--1350. IEEE, 2017.

\bibitem{fisac2019hierarchical}
Jaime~F Fisac, Eli Bronstein, Elis Stefansson, Dorsa Sadigh, S~Shankar Sastry,
  and Anca~D Dragan.
\newblock Hierarchical game-theoretic planning for autonomous vehicles.
\newblock In {\em 2019 International Conference on Robotics and Automation
  (ICRA)}, pages 9590--9596. IEEE, 2019.

\bibitem{genewein2015bounded}
Tim Genewein, Felix Leibfried, Jordi Grau-Moya, and Daniel~Alexander Braun.
\newblock Bounded rationality, abstraction, and hierarchical decision-making:
  An information-theoretic optimality principle.
\newblock {\em Frontiers in Robotics and AI}, 2:27, 2015.

\bibitem{gigerenzer2009homo}
Gerd Gigerenzer and Henry Brighton.
\newblock Homo heuristicus: Why biased minds make better inferences.
\newblock {\em Topics in cognitive science}, 1(1):107--143, 2009.

\bibitem{ingrand2017deliberation}
F{\'e}lix Ingrand and Malik Ghallab.
\newblock Deliberation for autonomous robots: A survey.
\newblock {\em Artificial Intelligence}, 247:10--44, 2017.

\bibitem{kappen2012optimal}
Hilbert~J Kappen, Vicen{\c{c}} G{\'o}mez, and Manfred Opper.
\newblock Optimal control as a graphical model inference problem.
\newblock {\em Machine learning}, 87(2):159--182, 2012.

\bibitem{lambert2020stein}
Alexander Lambert, Adam Fishman, Dieter Fox, Byron Boots, and Fabio Ramos.
\newblock Stein variational model predictive control.
\newblock {\em arXiv preprint arXiv:2011.07641}, 2020.

\bibitem{littman1994markov}
Michael~L Littman.
\newblock Markov games as a framework for multi-agent reinforcement learning.
\newblock In {\em Machine learning proceedings 1994}, pages 157--163. Elsevier,
  1994.

\bibitem{lutjens2019safe}
Bj{\"o}rn L{\"u}tjens, Michael Everett, and Jonathan~P How.
\newblock Safe reinforcement learning with model uncertainty estimates.
\newblock In {\em 2019 International Conference on Robotics and Automation
  (ICRA)}, pages 8662--8668. IEEE, 2019.

\bibitem{mellinger2011}
Daniel Mellinger and Vijay Kumar.
\newblock Minimum snap trajectory generation and control for quadrotors.
\newblock In {\em 2011 IEEE International Conference on Robotics and
  Automation}, pages 2520--2525, 2011.

\bibitem{ortega2013thermodynamics}
Pedro~A Ortega and Daniel~A Braun.
\newblock Thermodynamics as a theory of decision-making with
  information-processing costs.
\newblock {\em Proceedings of the Royal Society A: Mathematical, Physical and
  Engineering Sciences}, 469(2153):20120683, 2013.

\bibitem{ortega2015information}
Pedro~A Ortega, Daniel~A Braun, Justin Dyer, Kee-Eung Kim, and Naftali Tishby.
\newblock Information-theoretic bounded rationality.
\newblock {\em arXiv preprint arXiv:1512.06789}, 2015.

\bibitem{osborne2004introduction}
Martin~J Osborne et~al.
\newblock {\em An introduction to game theory}, volume~3.
\newblock Oxford university press New York, 2004.

\bibitem{pacelli2021robust}
Vincent Pacelli and Anirudha Majumdar.
\newblock Robust control under uncertainty via bounded rationality and
  differential privacy.
\newblock {\em arXiv preprint arXiv:2109.08262}, 2021.

\bibitem{crazyswarm}
James~A. Preiss*, Wolfgang H\"onig*, Gaurav~S. Sukhatme, and Nora Ayanian.
\newblock Crazyswarm: {A} large nano-quadcopter swarm.
\newblock In {\em {IEEE} International Conference on Robotics and Automation
  ({ICRA})}, pages 3299--3304. {IEEE}, 2017.
\newblock Software available at \url{https://github.com/USC-ACTLab/crazyswarm}.

\bibitem{reeves2012computing}
Daniel Reeves and Michael~P Wellman.
\newblock Computing best-response strategies in infinite games of incomplete
  information.
\newblock {\em arXiv preprint arXiv:1207.4171}, 2012.

\bibitem{schwarting2018planning}
Wilko Schwarting, Javier Alonso-Mora, and Daniela Rus.
\newblock Planning and decision-making for autonomous vehicles.
\newblock {\em Annual Review of Control, Robotics, and Autonomous Systems},
  1:187--210, 2018.

\bibitem{schwarting2021stochastic}
Wilko Schwarting, Alyssa Pierson, Sertac Karaman, and Daniela Rus.
\newblock Stochastic dynamic games in belief space.
\newblock {\em IEEE Transactions on Robotics}, 2021.

\bibitem{simon1955behavioral}
Herbert~A Simon.
\newblock A behavioral model of rational choice.
\newblock {\em The quarterly journal of economics}, 69(1):99--118, 1955.

\bibitem{spica2020real}
Riccardo Spica, Eric Cristofalo, Zijian Wang, Eduardo Montijano, and Mac
  Schwager.
\newblock A real-time game theoretic planner for autonomous two-player drone
  racing.
\newblock {\em IEEE Transactions on Robotics}, 36(5):1389--1403, 2020.

\bibitem{wang2019game}
Mingyu Wang, Zijian Wang, John Talbot, J~Christian Gerdes, and Mac Schwager.
\newblock Game theoretic planning for self-driving cars in competitive
  scenarios.
\newblock In {\em Robotics: Science and Systems}, 2019.

\bibitem{williams2018best}
Grady Williams, Brian Goldfain, Paul Drews, James~M Rehg, and Evangelos~A
  Theodorou.
\newblock Best response model predictive control for agile interactions between
  autonomous ground vehicles.
\newblock In {\em 2018 IEEE International Conference on Robotics and Automation
  (ICRA)}, pages 2403--2410. IEEE, 2018.

\bibitem{williams2017information}
Grady Williams, Nolan Wagener, Brian Goldfain, Paul Drews, James~M Rehg, Byron
  Boots, and Evangelos~A Theodorou.
\newblock Information theoretic mpc for model-based reinforcement learning.
\newblock In {\em 2017 IEEE International Conference on Robotics and Automation
  (ICRA)}, pages 1714--1721. IEEE, 2017.

\end{thebibliography}
}
\end{document}